\pdfoutput=1

\documentclass[11pt]{article}

\usepackage{ACL2023}

\usepackage{times}
\usepackage{latexsym}

\usepackage[T1]{fontenc}

\usepackage[utf8]{inputenc}

\usepackage{microtype}

\usepackage{inconsolata}

\usepackage{comment}
\usepackage{graphicx}
\usepackage[normalem]{ulem}
\usepackage{booktabs}
\usepackage{subfig}
\usepackage{caption}
\usepackage{multirow}

\title{Making a Long Story Short in Conversation Modeling}

\author{Yufei Tao \\
  Portland State University \\
  \texttt{yutao@pdx.edu} \\\And
  Tiernan Mines \\
  Hello Lamp Post \\
  \texttt{tiernan@hlp.city} \\\And
  Ameeta Agrawal \\
  Portland State University \\
  \texttt{ameeta@pdx.edu} \\}

\begin{document}
\maketitle

\begin{abstract}
Conversation systems accommodate diverse users with unique personalities and distinct writing styles.  Within the domain of multi-turn dialogue modeling, this work studies the impact of varied utterance  lengths on the quality of subsequent responses generated by conversation models. Using GPT-3 as the base model, multiple dialogue datasets, and several metrics, we conduct a thorough exploration of this aspect of conversational models. Our analysis sheds light on the complex relationship between utterance lengths and the quality of follow-up responses generated by dialogue systems. Empirical findings suggests that, for certain types of conversations, utterance lengths can be reduced by up to 72\% without any noticeable difference in the quality of follow-up responses.
\end{abstract}


\section{Introduction}


Recent research has made solid strides towards improving language models for dialogue applications and open-domain conversational agents \cite{shuster2022blenderbot,schulman2022chatgpt,thoppilan2022lamda,patil2023athena,wang2023ericson}. Numerous challenges associated with modeling multi-turn dialogues have been examined, with most prior work focused on expanding or restoring incomplete utterances \cite{su2019improving, inoue-etal-2022-enhance}.


\begin{figure} [t!]
\centering
\includegraphics[scale=0.3]{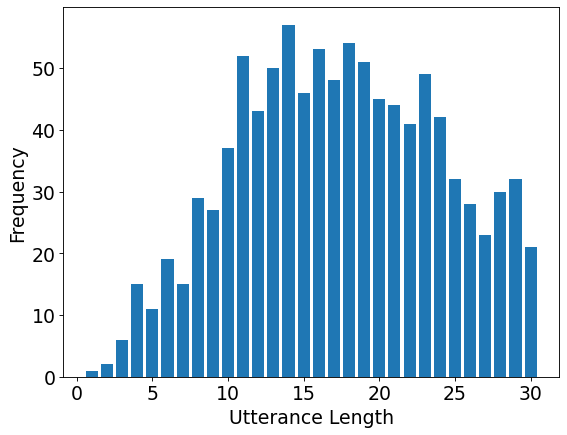}
\includegraphics[scale=0.3]{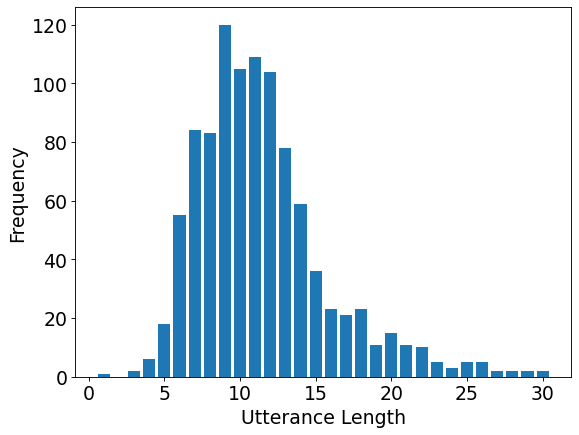}
\caption{Histograms showing the distribution of utterance lengths (words), as calculated from 1000 random samples from two datasets: (top) Topical-Chat  and (bottom) \textsc{ProsocialDialog}.} 
\label{fig:U2Length}
\end{figure}


An important feature of language production is the flexibility of lexical selection where speakers or writers choose specific words or lexical items to convey meaning in a given context \cite{jacobs2023chimpanzee}. This typically involves decisions regarding which words, phrases, or expressions to use to effectively communicate a message. Research indicates that vocabulary and grammatical structures are shaped by the context in which the utterance is produced, personal style, sociolinguistic factors (e.g., age), as well as discourse-level considerations \cite{bell1984language, bard2000controlling, tagg2014audience}. Consequently, the length of an utterance within a conversation exhibits a wide spectrum, ranging from succint expressions of just a few words to fully self-contained statements. To illustrate, Figure~\ref{fig:U2Length} presents the histogram plots of distribution of utterance lengths derived from two existing multi-turn conversation datasets showing a considerable variation.

Given such variation in our utterances, one natural question to ask is whether the length of our utterances influences the subsequent response, specifically the automatically generated response from a conversation model. This question becomes even more important when viewed through the lens of efficiency and inclusivity, particularly as access to cutting-edge conversation models becomes increasingly available primarily through paid services, often on a pay-per-token basis. In this work, we delve into the impact of utterance lengths on  conversation models' response generation, by modifying the length of the utterances as long or short, while keeping their essential meaning fairly unchanged.



Our empirical analysis considers five conversation datasets and several evaluation metrics, including both automatic and human evaluation. Interestingly, our findings  suggest that a substantial reduction in utterance length by almost 72\% results in as little as 8\% drop in METEOR score and 0.45\% drop in BERTScore. In other words, by reducing the number of tokens used as input, there emerges potential not only to reduce the computational costs of conversational systems, but also do so without any noticeable compromises in performance. 

\section{Related Work}


The context in which an utterance is produced heavily influences the choice of words and the grammatical structures \cite{jacobs2023chimpanzee}, and this is especially relevant in multi-turn dialogues where the length of utterances can vary widely. Most prior works in dialogue modeling have largely focused on expanding human utterances for contextual completeness by rewriting them,  and several models have been introduced for restoring incomplete utterances and including coreferred or omitted information to help multi-turn dialogue modeling \cite{liu2020incomplete,inoue-etal-2022-enhance}. However, these may result in unnecessary verbosity. 



Large language models such as GPT-3 \cite{brown2020language} and subsequent iterations such as GPT-4 and ChatGPT have garnered significant attention and adoption in the field of  conversation modeling \cite{tack2022ai, kumar2022exploring, abdelghani2022gpt,wang2023zeroshot, bdcc7030124, kalyan2023survey}. Their immense parameter sizes, reaching into the billions, enable them to capture intricate nuances in language and generate diverse and contextually relevant responses. However, it is worth noting that certain models\footnote{At the time of writing, some large language models can only be accessed via an API by paying a fee per some $n$ number of tokens (e.g., inferencing OpenAI's GPT-3 \texttt{davinci} models cost \$0.02 per 1K tokens).} come with considerable associated costs, often operating under the pay-as-you-go paradigm, where charges are typically computed based on the number of tokens utilized. 

Recent studies like FrugalGPT 
 \cite{chen2023frugalgpt} and LongLLMLingua 
 \cite{jiang2023longllmlingua} emphasize cost and performance optimization in LLMs. FrugalGPT explores cost-effective querying strategies, while LongLLMLingua focuses on prompt compression for efficiency in long context scenarios. 
 Our work complements these studies by specifically investigating the effect of {\em reducing} the utterance length on the model's performance in dialogue systems.

\begin{figure*} [t!]
\centering
\includegraphics[scale=1]{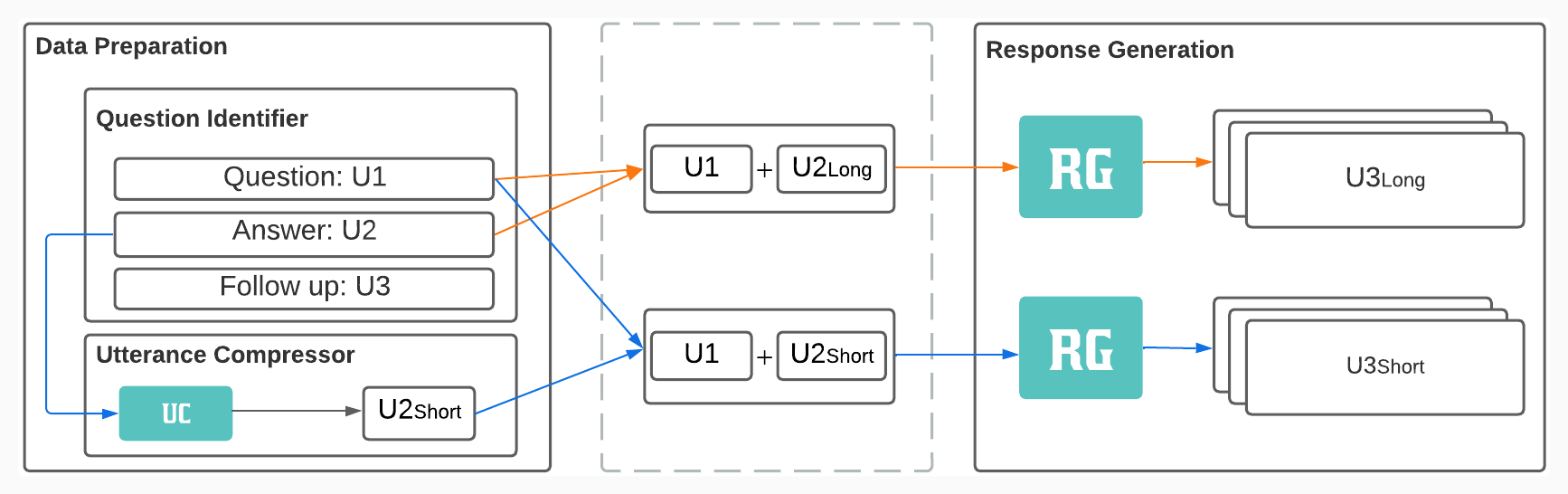}
\caption{Schematic illustration of the modeling process.} 
\label{fig:framework}
\end{figure*}

\begin{figure*} [t!]
\centering
\frame{\includegraphics[scale=0.6]{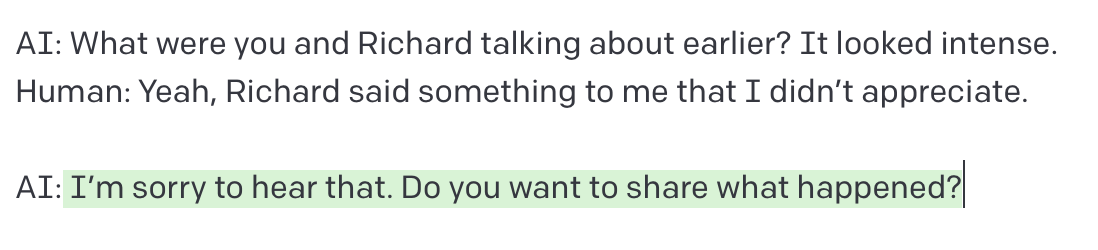}}
\caption{Example prompt simulating AI-Human conversation and a generated response (in green).} 
\label{fig:example_conversation}
\end{figure*}
\section{Model Description}

\subsection{Problem Formulation} Assume a conversation $\mathcal{C}=\{U_1, U_2, ..., U_n\}$ of $n$ utterances, where each utterance is a sequence of tokens $U_i=\{w_1,w_2,...,w_m\}$ of length $m$. We are concerned with specific subsets of a conversation consisting of three consecutive utterances ($U_1, U_2, U_3)$ where:
\begin{itemize}
    \item $U_1$ is a question or a query, 
    \item $U_2$ is a subsequent answer or response to $U_1$, and
    \item $U_3$ is a follow-up response to $U_2$.
\end{itemize}

We specifically focus on extracting  subsets of conversations where $U_1$ represents a question as questions inherently set the stage for informative and contextually connected responses. As such, this setup significantly increases the likelihood of $U_2$ and, consequently, $U_3$ being contextually relevant. Under this configuration, the goal is to investigate how the length of $U_2$ (either long or short) affects a model's follow-up response $U_3$. In other words, given $U_1$ along with a longer $U_{2_long}$ or a shorter $U_{2_{short}}$, we  generate and analyze the corresponding $U_{3_{long}}$ or $U_{3_{short}}$. Figure~\ref{fig:framework} presents an overview of the modeling process which includes two primary steps: data preparation and response generation.

\subsection{Data Preparation} This includes two sub-steps described as follows:

\medskip
\noindent (1) \textbf{Question Identifier} \quad  From a conversation we specifically select instances where $U_1$ is determined to be a question if it contains a question mark, ensuring that $U_1$ and $U_2$ are a question-answer pair, respectively, to maximize the contextual similarity between the two utterances and to minimize the possibility of topic shift.



\medskip
\noindent (2) \textbf{Utterance Compressor} \quad Next, we sample conversations where the length of $U_2$ is  more than some threshold $t_{long}$ to serve as our $U_{2_{long}}$ instances. Note that $U_{2_{long}}$ is the original unmodified utterance from the conversation. For reducing the length of these utterances to  shorter utterances while maintaining their overall meaning, one could employ a heuristically based approach or rewrite it automatically. We choose a model-in-the-loop module to generate $U_{2_{short}}$ from $U_{2_{long}}$ by prompting a generative language model as follows: 

\noindent \textcolor{gray}{
\texttt{ Q: Convert this sentence to another full sentence as short as it can be while keeping the same meaning, strongly prefer less than \{$t_{short}$\} words:} + \{$U_{2_{long}}$\}.} 


This prompt is used to generate shorter versions $U_{2_{short}}$ of the answer utterance. In our experiments, we use OpenAI's GPT-3 model and to ensure the validity of the generated condensed versions, we further manually reviewed each example and filtered out those that were not similar in meaning. As the focus of this study is to investigate the effect of utterance lengths, developing more efficient methods for compressing the utterances is left for future work. 



\subsection{Response Generation} Recall that $U_1$ is a question, $U_{2_{long}}$ is the longer/original response to $U_1$, and $U_{2_{short}}$ is the shorter response to $U_1$. The next step is to generate the follow-up responses $U_{3_{long}}$ and $U_{3_{short}}$, for $U_{2_{long}}$ and $U_{2_{short}}$, respectively. 


While several good conversation models exist, we generate these follow-up responses using GPT-3 by simulating a conversation between AI and a human. Our prompts are designed as follows:

\textcolor{gray}{\texttt{{\em AI: \qquad $\{U_1\}$}}}

\textcolor{gray}{\texttt{{\em Human:\quad $\{U_{2_{long/short}}\}$ }}}

\textcolor{gray}{\texttt{{\em AI: \qquad $\{U_{3_{long/short}}\}$ }}}

Following this design, we can facilitate the model to ask the question ($U_1$) first, which we then answer with $U_{2_{long}}$ and $U_{2_{short}}$, and finally collect the responses generated by the model. Figure~\ref{fig:example_conversation} presents an example prompt and output from GPT-3. 


\setlength\tabcolsep{20pt}
\begin{table*}[!h]
\centering
\begin{tabular}{lp{11.5cm}}
\toprule
 \textbf{Utterance} & \textbf{Text}\\
\midrule
${U}_1$ & {\em What were you and Richard talking about earlier? It looked intense.}  \\
 ${U}_{2_{long}}$ & {\em Yeah, Richard said something to me that I didn't appreciate.} \\
 ${U}_{2_{short}}$ & {\em Richard offended me.}\\
 ${U}_3$ & {\em Oh, no. I know how insensitive he can be. What has he done now?}\\
 ${U}_{3_{long}}$ & {\em I'm sorry to hear that. Can you tell me more about the situation?}
\\
 ${U}_{3_{short}}$ & {\em I'm sorry to hear that. Can you tell me what happened?}\\
\bottomrule
\end{tabular}
\caption{Sample instance from \textsc{TimeDial} dataset. $U_1$ denotes the question utterance,  $U_{2_{long}}$ is the original long response, $U_{2_{short}}$ is the condensed response, $U_3$ is the reference utterance from the dataset, and $U_{3_{long}}$ and $U_{3_{short}}$ are the model generated utterances.}
\label{tab:example}
\end{table*}

\subsection{Implementation}
The GPT-3 model we used is \texttt{text-davinci-003}, which was built on top of InstructGPT. For all the experiments, we used the same settings when calling the GPT-3 API as utterance compressor and response generator. The following hyperparameter settings were used: a sampling temperature of 0.9 to generate more diverse responses, a maximum number of generated tokens limited to 150, nucleus sampling set as default to 1 to choose the highest probability response, frequency of penalty set to 0 to not penalize frequently used words, presence penalty set to 0.6 to penalize words that appear frequently in the input text, and $n$ set as 3 to get the best three responses from GPT-3. Based on preliminary experiments, the length threshold $t_{long}$ is empirically set as 7 words and $t_{short}$ as 4 words.



\section{Experiment Setup}
This section describes the datasets and the evaluation metrics used in our analysis.

\subsection{Datasets}


Five existing conversation datasets are used, from which we extract subconversations consisting of three consecutive  utterances: $U_1$, $U_{2_{long}}$ and $U_3$. Note that $U_3$ serves as our reference text against which we evaluate the generated responses. One sample instance is shown in Table~\ref{tab:example}, while Table~\ref{tab:dataset_info} presents the statistics of all five datasets. The datasets include: 

\begin{itemize}
\item \textbf{\textsc{ProsocialDialog}} (PD) \cite{kim2022prosocialdialog}, a large-scale multi-turn dialogue dataset aimed at teaching conversational agents to respond to problematic content in accordance with social norms. The dataset covers topics that are unethical, problematic, biased, or toxic. 

\item \textbf{Commonsense-Dialogues} (CD) \cite{zhou2021commonsense}, a crowdsourced dataset of dialogues grounded in social contexts, which involve the utilization of commonsense. 

\item \textbf{\textsc{TimeDial}} (TD) \cite{qin2021timedial}, a crowdsourced dataset that contains multiple-choice cloze tasks. 

\item \textbf{Topical-Chat} (TC) \cite{Gopalakrishnan2019TopicalChat}, a dataset with human-human conversations about knowledge spanning eight broad topics (fashion, politics, books, sports, general entertainment, music, science and technology, and movies). 

\item \textbf{Ubuntu Dialogue} (UD) \cite{Lowe2015Ubuntu}, a dataset with two-person conversations extracted from the Ubuntu chat logs that  provide technical support for various Ubuntu-related problems. 
\end{itemize}

\setlength\tabcolsep{8pt}
\begin{table}[t!]
\centering
\begin{tabular}{lc}
\toprule
\textbf{Dataset} & \textbf{\# Conv.} \\
\midrule
\textsc{ProsocialDialog} (PD) & 636 \\
Commonsense-Dialogues (CD) & 490 \\
\textsc{TimeDial} (TD) & 533 \\
Topical-Chat (TC) & 579 \\
Ubuntu Dialogue (UD) & 567 \\
\bottomrule
\end{tabular}%
\caption{Statistics of the datasets. `\#Conv.' indicates the number of subconversations extracted and used in this work where $U_1$ is a question.}
\label{tab:dataset_info}
\end{table}

\setlength\tabcolsep{4.2pt}
\begin{table*}[h]
\centering
\begin{tabular}{c|cccc|cccc|cccc}
\toprule
 &
  \multicolumn{4}{c|}{\textbf{ROUGE-L}} &
  \multicolumn{4}{c|}{\textbf{METEOR}} &
  \multicolumn{4}{c}{\textbf{BERTScore}} \\ \midrule
 &
  \multicolumn{2}{c|}{Avg} &
  \multicolumn{2}{c|}{Max} &
  \multicolumn{2}{c|}{Avg} &
  \multicolumn{2}{c|}{Max} &
  \multicolumn{2}{c|}{Avg} &
  {Max} \\ \midrule
 &
  {L} &
 \multicolumn{1}{c|} {S} &
 {L} &
  \multicolumn{1}{c|}{S} &
  {L} &
 \multicolumn{1}{c|} {S} &
  {L} &
 \multicolumn{1}{c|} {S} &
  {L} &
 \multicolumn{1}{c|} {S} &
  {L} &
  {S} \\ \midrule

  \texttt{PD} &
  \multicolumn{1}{r}{{0.12}} &
  \multicolumn{1}{r|}{0.11} &
  \multicolumn{1}{r}{{0.16}} &
  0.15 &
  \multicolumn{1}{r}{{0.11}} &
  \multicolumn{1}{r|}{0.11} &
  \multicolumn{1}{r}{{0.15}} &
  0.14 &
  \multicolumn{1}{r}{{0.86}} &
  \multicolumn{1}{r|}{0.86} &
  \multicolumn{1}{r}{{0.87}} &
  0.87 \\ 
\texttt{CD} &
  \multicolumn{1}{r}{{0.14}} &
  \multicolumn{1}{r|}{0.12} &
  \multicolumn{1}{r}{{0.19}} &
  0.17 &
  \multicolumn{1}{r}{{0.12}} &
  \multicolumn{1}{r|}{0.11} &
  \multicolumn{1}{r}{{0.17}} &
  0.15 &
  \multicolumn{1}{r}{{0.87}} &
  \multicolumn{1}{r|}{0.87} &
  \multicolumn{1}{r}{{0.88}} &
  0.88 \\ 

\texttt{TD} &
  \multicolumn{1}{r}{{0.13}} &
  \multicolumn{1}{r|}{0.11} &
  \multicolumn{1}{r}{{0.17}} &
  0.15 &
  \multicolumn{1}{r}{{0.12}} &
  \multicolumn{1}{r|}{0.11} &
  \multicolumn{1}{r}{{0.17}} &
  0.15 &
  \multicolumn{1}{r}{{0.87}} &
  \multicolumn{1}{r|}{0.86} &
  \multicolumn{1}{r}{{0.88}} &
  0.87 \\ 
\texttt{TC} &
  \multicolumn{1}{r}{{0.12}} &
  \multicolumn{1}{r|}{0.11} &
  \multicolumn{1}{r}{{0.16}} &
  0.15 &
  \multicolumn{1}{r}{{0.12}} &
  \multicolumn{1}{r|}{0.11} &
  \multicolumn{1}{r}{{0.15}} &
  0.15 &
  \multicolumn{1}{r}{{0.85}} &
  \multicolumn{1}{r|}{0.85} &
  \multicolumn{1}{r}{{0.86}} &
  0.86 \\ 
\texttt{UD} &
  \multicolumn{1}{r}{{0.08}} &
  \multicolumn{1}{r|}{0.07} &
  \multicolumn{1}{r}{{0.12}} &
  0.09 &
  \multicolumn{1}{r}{{0.05}} &
  \multicolumn{1}{r|}{0.04} &
  \multicolumn{1}{r}{{0.08}} &
  0.06 &
  \multicolumn{1}{r}{{0.84}} &
  \multicolumn{1}{r|}{0.83} &
  \multicolumn{1}{r}{{0.84}} &
  0.84 \\ \midrule
\textbf{Avg.} &
  \multicolumn{1}{c}{{0.12}} &
  \multicolumn{1}{c|}{0.10} &
  \multicolumn{1}{c}{{0.16}} &
  \multicolumn{1}{c|}{0.14} &
  \multicolumn{1}{c}{{0.11}} &
  \multicolumn{1}{c|}{0.10} &
  \multicolumn{1}{c}{{0.14}} &
  \multicolumn{1}{c|}{0.13} &
  \multicolumn{1}{c}{{0.86}} &
  \multicolumn{1}{c|}{0.85} &
  \multicolumn{1}{c}{{0.87}} &
  \multicolumn{1}{c}{0.86} \\ 
  \midrule
\textbf{Diff. (L-S)} &
  \multicolumn{2}{c|}{0.02} &
  \multicolumn{2}{c|}{0.02} &
  \multicolumn{2}{c|}{0.01} &
  \multicolumn{2}{c|}{0.01} &
  \multicolumn{2}{c|}{0.01} &
  \multicolumn{2}{c}{0.01} \\ \bottomrule
\end{tabular}%
\caption{Experimental results of generating follow-up responses in conversations. `L' denotes the results with long form utterances, and, conversely, `S' denotes the results with shorter utterances. Due to the variability of responses, for each setting, we obtain three model generated responses. `Avg.' is the average of the three generated responses, whereas `Max.' reports their highest score. The datasets include PD (\textsc{ProsocialDialog}, CD (Commonsense-Dialogues), TD (\textsc{TimeDial}), TC (Topical-Chat), UD (Ubuntu Dialogue). The last row `Diff. (L-S)' presents the difference in the overall average scores of `L' and 'S'.}
\label{table:results}
\end{table*}

\subsection{Evaluation Metrics} 

We report the results using a variety of metrics of automatic evaluation as well as human assessment.


\medskip
\noindent \textbf{Automatic Evaluation} \quad To measure the quality of generated follow-up responses, we use three metrics to compare the similarity between $U_{3_{long/short}}$ and the reference response $U_3$. (i) \textbf{ROUGE-L\footnote{\url{https://pypi.org/project/rouge-score/}}} \cite{lin2004rouge} compares the longest common subsequence of words between the machine generated text and the reference text, normalized by the total number of words in the reference text. (ii) \textbf{METEOR\footnote{\url{https://www.nltk.org/api/nltk.translate.meteor_score.html}}} \cite{denkowski2014meteor} calculates the harmonic mean of unigram precision and recall, with a penalty for reordering of words and is a measure of how well the machine generated text aligns with the reference text.  (iii) \textbf{BERTScore\footnote{\url{https://huggingface.co/spaces/evaluate-metric/bertscore.html}}}
\cite{bert-score} uses the BERT model \cite{devlin2018bert} to evaluate the quality of machine generated text by calculating the similarity between the generated text and the reference text using cosine similarity between the embeddings. 

ROUGE-L measures overlap, considering word order and match length, while METEOR aligns generated text with reference text, and BERTScore assesses semantic similarity. All three metrics' scores range from 0 to 1, with 1 indicating a perfect match between the generated text and the reference text, and 0 indicating a complete mismatch.

\medskip
\noindent \textbf{Human Evaluation} \quad 
We also conduct a manual  assessment of the generated follow-up responses by having annotators estimate the similarity between the reference $U_3$ from the dataset and the generated responses $U_{3_{long}}$ and $U_{3_{short}}$. We randomly selected 8 samples from each of 5 datasets for a total of 40 evaluation samples.  Each sample contains $U_3$, $U_{3_{long}}$ and $U_{3_{short}}$. Four annotators were asked whether $U_{3_{long}}$ or $U_{3_{short}}$ is more similar to $U_3$ \textbf{( $U_{3_{long}}$ or $U_{3_{short}}$)}, whether both of them were equally similar (\textbf{both}), or whether neither of them was similar to $U_3$ (\textbf{neither}). A moderate level of inter-annotator agreement was found  (Fleiss’ Kappa = 0.58). 


\section{Results and Discussion}



\begin{table*}[h]
\centering
\setlength{\tabcolsep}{16pt}
\begin{tabular}{l|ccc|c|cc}
\toprule
& $U_{2_{long}}$ &	$U_{2_{short}}$	&\% condensing	& $U_3$ & $U_{3_{long}}$ &	$U_{3_{short}}$\\
\midrule
PD &	10.44 &	3.673& 64.8 &	17.98 & 86.37	&86.24\\
CD	&14.94	&4.01 	& 73.1 &	9.95&	48.37&	45.12\\
TD	&17.44	&4.60	& 73.5	&12.81	&55.13	&50.19\\
TC	&20.07	&5.52	& 72.4	&20.62	&93.66	&82.91\\
UD	&15.15	&3.83	& 74.7	&9.68 &	113.20	&124.31\\
\midrule
Avg.	&15.61 &	4.33 &	71.7 &	14.21	&79.35 &	77.76\\
\bottomrule
\end{tabular}
\caption{Comparison of length differences of $U_2$ and $U_3$ across five datasets. Even though there's a substantial 64-75\% compression from $U_{2_{long}}$ to $U_{2_{short}}$, the lengths of $U_{3_{long}}$ to $U_{3_{short}}$ remain consistently similar.}
\label{tab:lengths}
\end{table*}



From the results detailed in Table~\ref{table:results}, we observe that, surprisingly, the average scores for the long and shorter length settings  remain comparable, with the difference between them (as indicated in the last row) ranging from 0.01 to 0.02. These findings suggest that, while using the longer $U_{2_{long}}$ input yields a slightly better quality in the generated $U_{3_{long}}$  compared to using $U_{2_{short}}$  for generating $U_{3_{short}}$, the actual difference between the two versions of the generated texts remains minimal (around 1\% for ROUGE-L and METEOR, and 0.4\% for BERTScore).




Next, we discuss the results of human evaluation. 54\% of the annotations were marked as `both' or `neither', whereas 22.5\% and 23\% of the annotations preferred $U_{3_{long}}$ and $U_{3_{short}}$, respectively, as the better response. This further confirms that the quality of $U_{3_{long}}$ and $U_{3_{short}}$ remains comparable as per human evaluation.


One possible explanation for the relatively small disparity in the quality between $U_{3_{long}}$ and $U_{3_{short}}$ is provided by further analysis of these responses. As Table~\ref{tab:lengths} illustrates, despite the significant compression of $U_{2_{long}}$ to $U_{2_{short}}$ by approximately 72\% (as indicated by `\% compressed'), the lengths of the generated responses $U_{3_{long}}$ and $U_{3_{short}}$ remain remarkably comparable, with differences not exceeding 2 words on average. Lastly, we notice that the GPT-3 model tends to generate responses that are substantially more verbose than $U_{3}$, an observation that aligns with findings reported in several recent works  \cite{goyal2023news,chiesurin2023dangers}.





These findings suggest that a significant reduction in the number of input tokens in these question-answer subconversations may not necessarily impact the generation of the follow-up response. This may be due to the presence of $U_1$ in the input which provides sufficient context for the model to generate the subsequent responses.





\section{Conclusion}

In this study, we explored the nuanced dynamics of utterance length in conversational modeling. Our investigation revealed that, particularly in question-answer and follow-up response contexts, significantly shorter utterances do not adversely impact the model's ability in generating coherent and contextually appropriate follow-up responses. 

The findings of this study suggest a potential avenue for exploring utterance length as a factor in enhancing the efficiency of language models for conversational tasks from a novel perspective. By acknowledging the effectiveness of shorter inputs, future research can examine alternative token reduction techniques and the linguistic nuances of shortened inputs, aiming to optimize the balance between brevity and performance.



\section*{Limitations}
This work has a few notable limitations. First, we measured the quality of the generated texts ($U_{3_{long/short}}$) by comparing them to the original dialogue utterance ($U_{3}$) as reference that was present in the dataset. However, in open-ended text generation, there can be several acceptable references. While our evaluation method captures essential aspects of the conversation, it might not cover every nuance. Recent LLM-based evaluations like G-Eval \cite{liu2023geval} which employs chain-of-thoughts or MEEP \cite{ferron_etal_2023_meep} which focuses on estimating dialogue engagingness could offer deeper insights into the quality of the generated responses. Additionally, the original average length of $U_3$ was found to be substantially shorter than the responses generated by the LLM, which could further impact the evaluation scores. It may be worth experimenting with setting GPT-3's maximum token limit closer to the average length of $U_3$. It is also worth mentioning that our empirical analysis focuses on utterances which are preceded by a question, therefore, making the response somewhat less unexpected. The effectiveness of this approach in conversations with sudden topic drifts or changes remains to be studied. We also acknowledge that compressing $U_2$ using GPT-3 may not be the most efficient approach and a heuristic method would be more ideal for this experiment considering the efficiency factor.  

Furthermore, this study was conducted with GPT-3, and since then, there have been significant advancements in the field of large language models, including the release of GPT-4 and other open-source models. Future work could benefit from replicating and extending this experiment with these advanced models to compare the effectiveness and efficiency of dialogue generation and compression.






\section*{Ethics Statement}
We acknowledge that in conversation datasets of natural language, potential toxic data instances may exist, which may further negatively propagate throughout the modeling process. During the compressing of $U_{2_{long}}$, it is possible that some utterances may become ambiguous or assume unintentional modified meaning. 

\section*{Acknowledgements}
We would like to thank the anonymous reviewers and the members of the PortNLP group for their insightful feedback.  This research was supported by the NSF under grant number SAI-P-2228783 and CRII:RI-2246174.



\bibliography{anthology,custom}
\bibliographystyle{acl_natbib}
\appendix


\setcounter{table}{0}
\renewcommand{\thetable}{A\arabic{table}}

\newpage

\end{document}